\documentclass[letterpaper, 10pt, twocolumn]{article}

\usepackage[a4paper, left=0.8in, right=0.8in, top=1in, bottom=1in, columnsep=0.4in]{geometry} 
\usepackage{amsmath, bm} 
\usepackage{graphicx} 
\usepackage{dcolumn} 
\usepackage{booktabs} 
\usepackage{multirow} 
\usepackage{subcaption} 
\usepackage{caption} 
\usepackage{titlesec} 
\usepackage{fancyhdr} 
\usepackage{libertine} 
\usepackage{hyperref} 
\usepackage{tcolorbox} 
\usepackage{enumitem} 
\usepackage{amsfonts} 
\usepackage[percent]{overpic}

\hypersetup{
    colorlinks=true,
    linkcolor=black,
    urlcolor=black,
    citecolor=black
}

\titleformat{\section}{\Large\bfseries\rmfamily}{\thesection}{1em}{}
\titleformat{\subsection}{\large\bfseries\rmfamily}{\thesubsection}{1em}{}
\titlespacing*{\section}{0pt}{1.5em}{1em} 
\titlespacing*{\subsection}{0pt}{1em}{0.5em} 

\title{\textbf{Optimizing 3D Gaussian Splatting via Point Cloud Upsampling}}

\author{
    Adrian Ramlal$^{1}$ \\
    Yan Song Hu$^{1}$, John S. Zelek$^{1}$ \\
    $^1$Vision and Image Processing Group, Systems Design Engineering, University of Waterloo\\
    \texttt{\{avdramla, y324hu, jzelek\}@uwaterloo.ca}
}
\date{}

\begin{document}

\maketitle


\begin{tcolorbox}[colback=gray!10, sharp corners, boxrule=0pt, left=5pt, right=5pt, top=5pt, bottom=5pt]
\begin{abstract}
3D Gaussian Splatting (3DGS) is a technique for creating and rendering 3D scenes, however its performance depends heavily on the quality of initial seed points. To improve 3DGS initialization, this study presents and evaluates several point cloud upsampling approaches: linear interpolation, triangular interpolation, spline-based surface reconstruction, moving least squares surface fitting, and Voronoi-based point generation. Additionally, this research introduces a depth-guided point lifting method that leverages depth maps to maintain geometric consistency with Structure-from-Motion (SfM) reconstructions. Through extensive experiments on the Mip-NeRF360 and Replica datasets, the proposed methods demonstrate improvements in reconstruction quality across diverse scene types. Results indicate that different upsampling strategies excel in different scenarios: surface reconstruction methods perform better with organic, detailed scenes, while simpler interpolation approaches are more suited for scenes dominated by piecewise-smooth geometries. In comparison, the depth-guided approach shows promise for adding geometry-aware points across the entire scene, importantly in texture-less regions. These findings, which provide preliminary practical guidelines for selecting appropriate upsampling methods based on scene characteristics and computational constraints, advances the understanding of how point cloud initialization affects 3DGS quality.
\end{abstract}
\end{tcolorbox}

\section{Introduction}





3D Gaussian Splatting (3DGS) \cite{kerbl3Dgaussians} has emerged as a groundbreaking rendering technique in computer vision and graphics. 3DGS represents scenes as a collection of Gaussian primitives, which are 3D semi-transparent clouds that vary in color and density. By strategically positioning and blending these Gaussian primitives, 3DGS can generate photorealistic representations of complex environments and synthesize novel viewpoints. As data-hungry spatial AI techniques continue to evolve, 3DGS has gained prominence as a powerful tool for tasks such as VR/AR, autonomous navigation, and scene understanding. The ability of 3DGS to provide dense volumetric representations of environments and enable the synthesis of novel viewpoints make it a compelling solution for spatial computing applications.

The generation of 3DGS representations from images requires two additional inputs: the relative poses of the input images and 3D points that serve as seed locations for the initial Gaussians. While it is established that 3DGS performance has a strong correlation with pose accuracy, the optimal strategy for initializing seed points remains an open research question. Traditional Structure-from-Motion (SfM) pipelines generate sparse point clouds that, while sufficient for camera pose estimation, may not provide optimal initialization for 3DGS. This sparsity can lead to gaps in the reconstructed scene and require additional optimization iterations to achieve satisfactory results. Moreover, the distribution of these points often follows high-texture regions, leaving important but texture-less areas under-represented.

This paper presents an investigation of methods for augmenting the initial point cloud used for seeding 3DGS, with the goal of improving scene reconstruction quality. Our results demonstrate that thoughtful augmentation of the initial point cloud can lead to improvements in the final reconstruction quality of 3DGS. Our primary contributions include:
\begin{enumerate}
    \item An evaluation of the impact of interpolation, surface reconstruction and point lifting upsampling methods on 3DGS reconstruction quality.
    \item Empirically derived guidelines for selecting appropriate upsampling methods based on scene characteristics.
\end{enumerate}

\section{Related Works}
Proper initialization of seed Gaussian positions is essential for the effectiveness of 3DGS. This is because the Gaussian cloning and pruning process in 3DGS is driven by heuristics calculated from existing Gaussians. Gaussian creation depends on cloning existing Gaussians, making it challenging to distribute them effectively if initial seeds are poorly placed. Furthermore, a recent work by Lu et al. \cite{lu2024poisonsplatcomputationcostattack} indicates that 3DGS does not impose penalties for excessive Gaussian numbers, so minimizing the initialization of unnecessary Gaussians can enhance computational efficiency and reduce memory usage. The original 3DGS paper by Kerbl et al. \cite{kerbl3Dgaussians} underscores this point, with an ablation study showing a decrease in PSNR performance of around 6 dB when Gaussians are initialized randomly. 

The original 3DGS paper used the point cloud output from the SfM system COLMAP \cite{schoenberger2016sfm} as seed points. However, COLMAP’s point cloud is not optimized as an initialization for 3DGS; rather, it is a byproduct of the pose estimation process. One straightforward approach to obtaining better seed points is to enhance the images with data from an additional sensor, such as LiDAR \cite{lim2024lidar3dgslidarreinforced3d} or depth sensors \cite{ha2024rgbdgsicpslam}. However, introducing an extra sensor is often not ideal. 

Machine-learned monocular depth enables point cloud augmentation with additional points without requiring extra sensor data. This approach is utilized by InstantSplat by Fan et al. \cite{fan2024instantsplat} to allow 3DGS training that finishes training in seconds. InstantSplat achieves this speed by leveraging Dust3R \cite{wang2024dust3rgeometric3dvision}, a machine learning-based SfM system that calculates depth to generate dense output maps. The dense and detailed initial point cloud provided by Dust3R is a key factor enabling the speed of the InstantSplat process. Another way to use machine learned depth information is proposed by Chan et al. \cite{10.1145/3664647.3681454}. Instead of using a different SfM system, the work by Chan et al. augments and realigns COLMAP point clouds using a Dense Prediction Transformer. 

While machine-learned augmentation is one approach to enhancing point cloud initialization, this paper focuses on improving seed point clouds using traditional computer vision and graphics methods. One method using traditional methods is by Seibt et al. \cite{10.2312:egp.20241038}, who propose improving the initial point cloud by detecting and tracking additional points found using a dense SfM system called DFM4SFM. A similar work is by Hu et al. \cite{hu2024realtimegaussiansplattingaccelerating} uses the Photometric SLAM system Direct Sparse Odometry instead of COLMAP to do more dense point tracking. In contrast, our approach augments the existing COLMAP-generated point clouds rather than replacing the SfM system altogether.

\section{Methods}


We present several approaches for upsampling sparse point clouds to generate augmented initialization seeds for 3DGS. Our methods encompass three distinct strategies: interpolation between existing points, local surface reconstruction for generating geometrically-consistent new points, and a depth-guided point lifting approach that leverages depth data. The input to our pipeline consists of sparse point clouds generated through COLMAP reconstruction \cite{schoenberger2016sfm}. Our interpolation and surface-fitting methods operate exclusively on the original reconstructed points to avoid potential cascading errors that could arise from iterative point generation. The final point cloud used for 3DGS initialization is the union of the original and generated points.

\subsection{Point Interpolation}

\subsubsection{Linear Upsampling}
Our first approach generates new points by implementing a weighted linear interpolation between existing points. For each new point, we randomly select a point $P_1$ and find its nearest neighbor $P_2$, using $k$-nearest neighbor search (kNN). Then, a randomly selected weight, $\alpha$, is utilized to interpolate the spatial coordinates (XYZ) and color (RGB):
\begin{equation}
P_{new} = \alpha P_1 + (1-\alpha)P_2
    \label{eq1}
    \end{equation}
Where $\alpha \in \left[0,1\right]$. This method maintains local density features while filling gaps between existing points.

\subsubsection{Triangular Upsampling}
Building upon the linear approach, triangular upsampling utilizes barycentric coordinates to generate points within the triangular regions formed by existing point triplets. For each new point, we select a random point $P_1$ and find its two $k$-nearest neighbors, $P_2$ and $P_3$, to form a triangle. Then we generate the point's coordinates and color using barycentric coordinates:
\begin{equation}
    P_{new} = \alpha P_1 + \beta P_2 + \gamma P_3
    \label{eq2}
\end{equation}
Where $\alpha + \beta + \gamma = 1$ and $\alpha, \beta, \gamma \geq 0$. This technique improves upon linear upsampling by considering the local surface geometry implied by the increased spatial context of point triplets.

\subsection{Surface Reconstruction}

\subsubsection{Spline Upsampling}
Our third approach leverages local B-spline surface fitting to generate new points that better preserve the underlying geometric structure. For each new point, we randomly select an existing point $P_c$ and identify its nearest neighbors ($k=25$) to form a local neighborhood $\mathcal{N}(P_c)$. Within this neighborhood, we fit a bi-variate B-spline surface $S(x,y)$ using:
\begin{equation}
    S(x,y) = \sum_{i=0}^{n_x} \sum_{j=0}^{n_y} c_{ij}N_{i,p}(x)N_{j,p}(y)
    \label{eq3}
\end{equation}
Where $N_{i,p}(x)$ and $N_{j,p}(y)$ are B-spline basis functions of degree $p=3$, $c_{ij}$ are control points determined through smoothing spline optimization ($s=0.1k$) and $n_x=n_y=\min(\max(4,\lfloor k/4 \rfloor),8)$ are the number of knots in each direction \cite{10.5555/265261}. To generate a new point, $P_{new}$, we randomly select an $x_i$ and $y_i$ value within $\mathcal{N}(P_c)$ and evaluate $S(x_i,y_i)$. RGB values for new points are computed through distance-weighted interpolation of the neighborhood values. This method captures surface geometries and generates smoothly distributed points by incorporating higher-order geometric information from the local neighborhood. 

\subsubsection{Moving Least Squares (MLS) Upsampling}
Our fourth approach employs MLS surface fitting to generate new points that accurately represent local surface geometry. For each new point, we randomly select an existing point $P_c$ and identify its nearest neighbors ($k=10$) to form a local neighborhood $\mathcal{N}(P_c)$. Within this neighborhood, we fit a local polynomial surface of degree $d=2$ using:
\begin{equation}
    f(x,y) = \sum_{i+j\leq d} a_{ij}x^iy^j
    \label{eq4}
\end{equation}
Where $a_{ij}$ are coefficients determined by solving the weighted least squares problem:
\begin{equation}
    \min_{a_{ij}} \sum_{p \in \mathcal{N}(P_c)} w(p)(z_p - f(x_p,y_p))^2
    \label{eq5}
\end{equation}
Where weights $w(p)$ inversely proportional to the distance from $P_c$ so that closer points contribute more heavily to the fit \cite{Fasshauer2003}. A new point, $P_{new}$, is generated by randomly sampling a position within the neighborhood region and projecting this position onto the fitted surface. The RGB value is interpolated from neighboring points using distance-weighted averaging. This method leverages a continuous surface model that adapts to local geometry to smooth point distributions and minimize distortions, especially in higher curvature regions, by capturing finer surface details.

\subsubsection{Voronoi Upsampling}
Our fifth approach utilizes Voronoi cell decomposition to generate new points that adapt to local point density \cite{chang2008surface}. For the original point cloud, we construct a Voronoi diagram $\mathcal{V}$ that partitions the space into cells, where each cell $V_i$ contains all points closer to the generating point $P_i$ than to any other point. To ensure bounded cells, we augment the point set with a bounding box extending 10\% beyond the point cloud extent. For each Voronoi cell $V_i$, we compute its properties:
\begin{equation}
\left\{
    \begin{aligned}
         & c_i = \frac{1}{|V_i|} \sum_{v \in V_i} v \quad \text{(cell center)}\\
         & r_i = \max_{v \in V_i} |v - c_i| \quad \text{(maximum radius)} \\
         & vol_i = \mathbb{E}[|v - c_i|]^3 \quad \text{(approximate volume)} \\
    \end{aligned}
\right.
\label{eq6} 
\end{equation}

New points are generated preferentially in larger Voronoi cells, which represent sparsely sampled regions. For each new point, we select a cell $V_i$ with probability proportional to its volume $vol_i$ and generate a random point $P_{new}$ within a sphere of radius $0.5r_i$ centered at $c_i$. The new point's attributes are interpolated from the $k=5$ nearest neighbors. This method balances the point distribution by generating more points in sparse regions while preserving local geometric features through attribute interpolation. 

\subsection{Depth-Guided Point Lifting}
Motivated by work from Chan et al. \cite{chan2024point}, we present a depth-guided point lifting approach that augments sparse SfM reconstructions while maintaining geometric accuracy. Our method synthesizes information from multiple sources: RGB images, their depth maps, and camera parameters estimated through COLMAP's SfM reconstruction \cite{schoenberger2016sfm}. For each image in the dataset, we transform 2D pixels into 3D points:

\begin{enumerate}
    \item Scale Factor Computation: To ensure geometric consistency with original SfM points, we compute optimal scale factor ($s$) by minimizing the error between projected depth points and their corresponding SfM points: 
    \begin{equation}
        s = \frac{\sum (p_{raw} \cdot (p_{orig} - c))}{\sum p_{raw}^2}
        \label{eq7}
    \end{equation} 
    Where $p_{raw}$ are the raw projected points, $p_{orig}$ are the original SfM points, and $c$ is the camera center.
    \item Point Generation: New 3D points are generated through the following steps:
    \begin{enumerate}
    \item Point Selection: We sample points from the central region of each image (50\% of image dimensions) to ensure reliable depth measurements.
    \item Coordinate Transformation: For each randomly selected pixel $p=(u,v)$, we convert it to world coordinates:
    \begin{equation}
        p_w = t_{inv} + (R_{inv} \cdot K^{-1}[u,v,1]^T)
        \label{eq8}
    \end{equation}
    Where $K$ is the intrinsic camera matrix.
    \item Depth Integration: We compute the final 3D position by scaling the ray from camera center to projected point using the depth value and optimal scale factors:
        \begin{equation}
            P_{3D} = c_{world} + s \cdot (d \cdot \hat{v})
            \label{eq9}
        \end{equation}
    Where $c_{world}$ is the camera center in world coordinates, $d$ is the depth value, and $\hat{v}$ is the unit vector from camera center to projected point.
    \end{enumerate}
    \item Color Assignment: RGB values are sampled from the corresponding pixels in the source image, to maintain visual consistency with the input data.
\end{enumerate}

This method combines geometric information from both SfM and depth maps, resulting in dense point clouds that maintain consistency with the original sparse reconstruction while adding detail in previously under-sampled regions. The use of a scale factor derived from existing SfM points ensures that new points are aligned with the global reconstruction, while the central region sampling strategy helps minimize the impact of depth estimation errors that are emphasized near image boundaries.

\section{Results and Discussion}
\begin{table*}[th]
    \caption{Upsampling Methods PSNR($\uparrow$) on Mip-NeRF360 Dataset \cite{barron2022mipnerf360}.}
    \centering
    \scalebox{0.72}{
        \begin{tabular}{l|c|ccc|ccc|ccc|ccc|ccc}
            \toprule 
            Scene & 3DGS & \multicolumn{3}{c|}{Linear} & \multicolumn{3}{c|}{Triangle} & \multicolumn{3}{c|}{Spline} & \multicolumn{3}{c|}{MLS} & \multicolumn{3}{c}{Voronoi}\\
            & & 4x & 8x & 16x & 4x & 8x & 16x & 4x & 8x & 16x & 4x & 8x & 16x & 4x & 8x & 16x \\
            \midrule
            bicycle & 25.106 & 25.270 & 25.261 & 25.225 & 25.248 & 25.262 & 25.236 & 25.163 & 25.160 & 25.188 & 25.227 & \textbf{25.285} & 25.281 & 25.143 & 25.120 & 25.182 \\
            bonsai & 32.291 & 32.275 & \textbf{32.660} & 32.262 & 32.413 & 32.406 & 32.532 & 32.250 & 32.520 & 32.620 & 32.487 & 32.572 & 32.554 & 32.534 & 32.629 & 32.629 \\
            counter & 29.112 & 29.228 & 29.280 & 29.285 & 29.292 & \textbf{29.353} & 29.301 & 29.199 & 29.284 & 29.334 & 29.226 & 29.310 & 29.322 & 29.194 & 29.259 & 29.227 \\
            flowers & 21.382 & 21.455 & 21.507 & 21.426 & 21.490 & 21.496 & 21.445 & 21.469 & 21.462 & 21.537 & 21.520 & 21.608 & \textbf{21.671} & 21.437 & 21.514 & 21.547 \\
            garden & 27.329 & 27.416 & 27.399 & 27.398 & 27.416 & 27.390 & 27.430 & 27.338 & 27.410 & 27.386 & 27.382 & 27.402 & \textbf{27.485} & 27.288 & 27.328 & 27.354 \\
            kitchen & 31.559 & 31.719 & 31.525 & 31.777 & 31.796 & 31.812 & 31.534 & 31.442 & 31.755 & 31.294 & \textbf{31.874} & 31.626 & 31.852 & 31.625 & 31.596 & 31.171 \\
            room & 31.620 & 31.452 & 31.599 & 31.634 & 31.740 & 31.596 & 31.789 & 31.879 & 31.733 & \textbf{31.983} & 31.865 & 31.873 & 31.915 & 31.772 & 31.878 & 31.951 \\
            stump & 26.616 & 26.819 & 26.838 & 26.841 & 26.814 & 26.858 & 26.832 & 26.706 & 26.751 & 26.775 & 26.835 & 26.844 & \textbf{26.864} & 26.852 & 26.773 & 26.710 \\
            treehill & 22.531 & 22.493 & 22.468 & 22.441 & 22.439 & 22.529 & 22.458 & 22.518 & 22.504 & 22.485 & 22.562 & 22.483 & 22.417 & \textbf{22.689} & 22.629 & 22.573 \\
            \bottomrule
        \end{tabular}
    }
    \label{tab1}
\end{table*}

Following 3DGS \cite{kerbl3Dgaussians}, we utilize the Mip-NeRF360 dataset \cite{barron2022mipnerf360} to evaluate our interpolation and reconstruction upsampling methods and we report the novel view synthesis metric of PSNR. Additionally, we evaluate on 8 scenes of the Replica Dataset \cite{replica19arxiv}, which provides synthetic depth maps alongside RGB images.

\subsection{Evaluation on Mip-NeRF360}
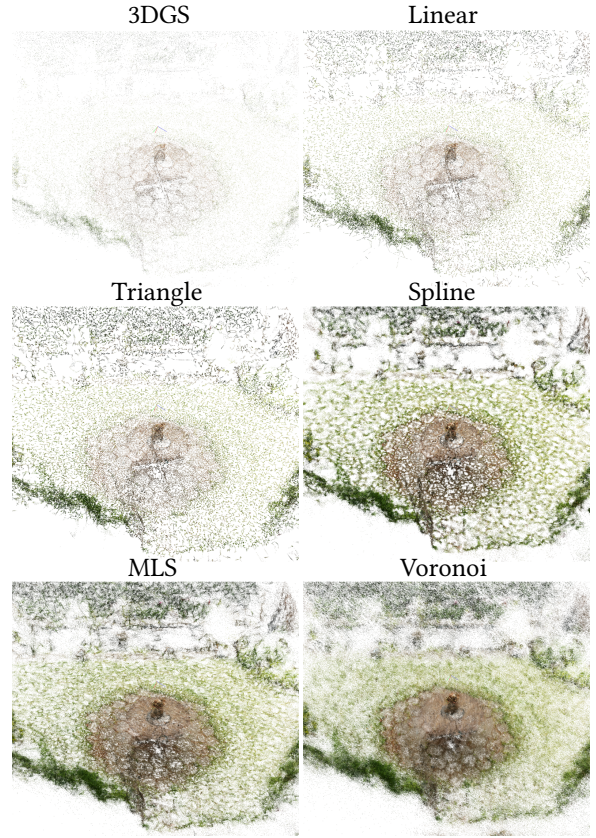
\begin{figure}[t]
    \centering
    \begin{overpic}[width=0.45\textwidth]{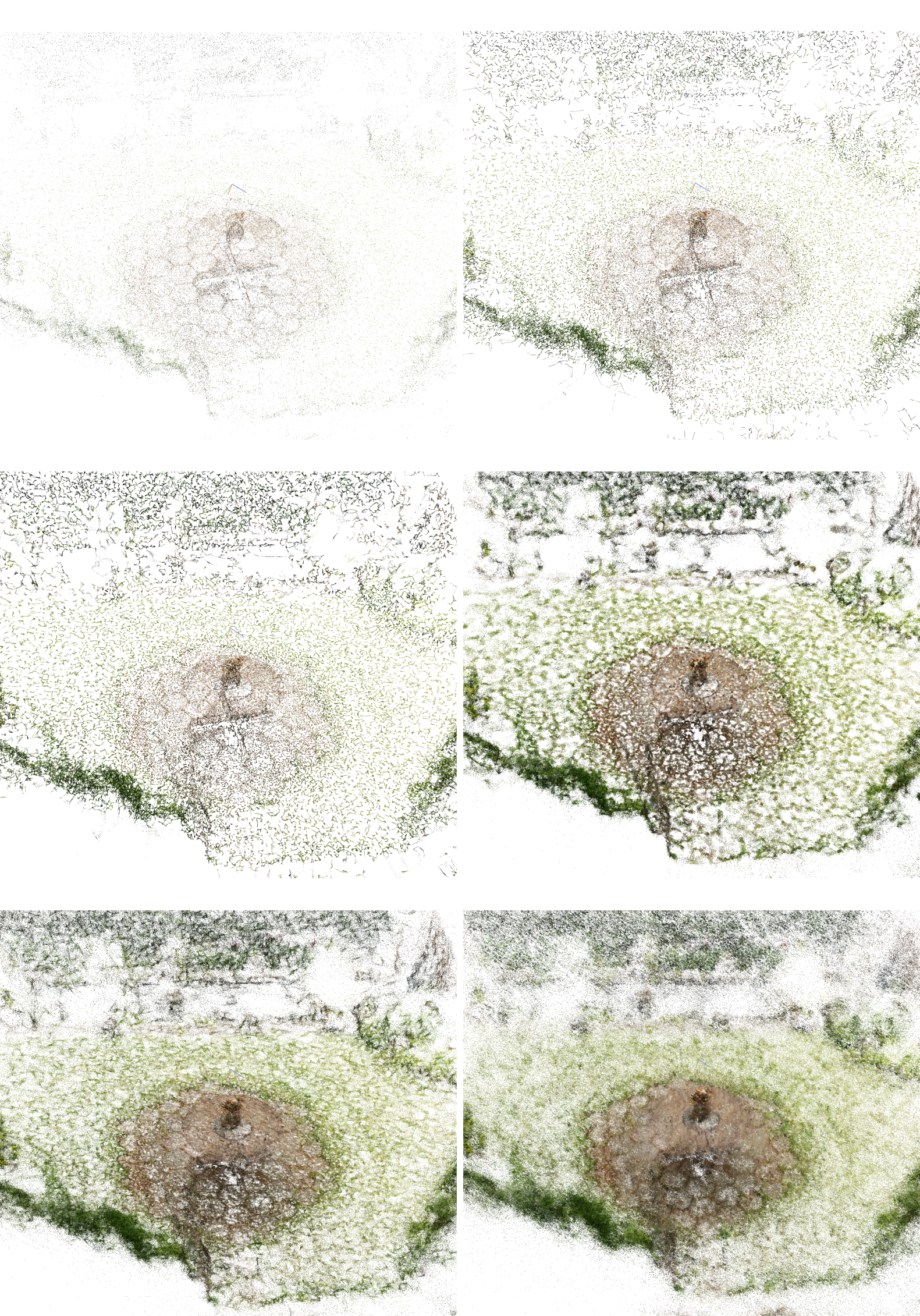}
        \put (14,98.5) {3DGS}
        \put (48,98.5) {Linear}
        \put (12,65) {Triangle}
        \put (48,65) {Spline}
        \put (14,31.5) {MLS}
        \put (47,31.5) {Voronoi}
    \end{overpic}
    \caption{The 16x upscaled point clouds of \textit{garden} from Mip-NeRF360 \cite{barron2022mipnerf360}.}
    \label{fig1}
\end{figure}
We evaluate on Mip-NeRF360 \cite{barron2022mipnerf360} across different upsampling ratios (4x, 8x, and 16x). Points clouds produced by our methods are shown in Fig.~\ref{fig1}. The results, presented in Table~\ref{tab1}, reveal several key insights into the effectiveness of these methods for 3DGS initialization.

Across all scenes, our upsampling methods generally maintain or improve the PSNR compared to the baseline 3DGS results with an average increase of approximately 0.258 dB per scene. The MLS approach consistently demonstrates superior performance, achieving the highest PSNR in 5 out of 8 scenes at various upsampling ratios. For instance, in the \textit{kitchen} scene, MLS achieves a PSNR of 31.874 at 4x upsampling, representing a notable improvement over the baseline PSNR of 31.559. Interestingly, the simpler Linear and Triangle interpolation methods also show promise, particularly at lower upsampling ratios. For instance, the Linear method at 8x upsampling achieves a PSNR of 32.660, an improvement of 0.369 dB, on the \textit{bonsai} scene.

\subsubsection{Guidelines for Method Selection}
Using the effectiveness of different upsampling methods across scenes, we derive guidelines for method selection based on scene characteristics, such as structural complexity and the presence of organic or planar forms. Firstly, in highly detailed scenes like \textit{counter} and \textit{room} surface reconstruction methods (e.g., Spline and MLS) tend to outperform simpler interpolation approaches, likely due to their ability to better capture subtle curvature and local detail. In scenes characterized by more organic content, like \textit{flowers} and \textit{bicycle}, MLS continues to show robust performance, suggesting its adaptability to irregular, natural shapes where preserving small-scale geometry is important. In contrast, for \textit{treehill}, a more planar environment, the Voronoi method exhibits higher PSNR that all other methods, to a greater extent at lower upsampling ratios. This suggests that Voronoi's adaptive point distribution strategy is effective at handling flat, low-texture surfaces. The performance of the interpolation methods on the \textit{bonsai} and \textit{counter} scenes is a consequence of the prevalence of piecewise-smooth geometries, resulting from many object of regular geometries, tend to be over-smoothed by the surface reconstruction methods.

Thus, our findings demonstrate that thoughtful point cloud upsampling can enhance 3DGS reconstruction quality. The choice between interpolation and surface reconstruction methods should consider both scene characteristics and computational constraints. Simpler methods can sometimes achieve comparable results with lower computational overhead and lower upsampling ratios (4x or 8x) often provide a favorable trade-off between point cloud density and reconstruction quality.

\subsection{Evaluation on Replica}

\begin{table*}[t]
    \caption{Upsampling Methods PSNR($\uparrow$) on Replica \cite{replica19arxiv}.}
    \centering
    \scalebox{0.72}{
        \begin{tabular}{l|c|cc|cc|cc|cc|cc|cc}
            \toprule 
            Scene & 3DGS & \multicolumn{2}{c|}{Linear} & \multicolumn{2}{c|}{Triangle} & \multicolumn{2}{c|}{Spline} & \multicolumn{2}{c|}{MLS} & \multicolumn{2}{c|}{Voronoi} & \multicolumn{2}{c}{Depth}\\
            & & 16x & 32x & 16x & 32x & 16x & 32x & 16x & 32x & 16x & 32x & 16x & 32x \\
            \midrule
            office0 & 43.505 & 44.001 & 44.177 & 44.089 & 44.204 & 43.967 & 44.039 & 44.173 & \textbf{44.373} & 44.067 & 44.168 & 44.177 & 44.281 \\
            office1 & 41.126 & 41.117 & 41.205 & 41.046 & 41.301 & 41.284 & 41.377 & 41.538 & 41.689 & 41.280 & 41.437 & 41.812 & \textbf{42.029} \\
            office2 & 37.941 & 38.099 & 38.035 & 38.136 & 38.128 & 38.203 & 38.223 & 38.242 & 38.266 & 38.148 & 38.305 & 38.393 & \textbf{38.406} \\
            office3 & 37.495 & 37.852 & 37.881 & 37.892 & 37.837 & 37.970 & 38.087 & 38.041 & \textbf{38.171} & 37.899 & 38.002 & 37.728 & 37.888 \\
            office4 & 38.902 & 39.186 & 38.962 & 38.968 & 39.131 & 38.912 & 39.032 & 39.288 & \textbf{39.534} & 39.244 & 39.309 & 39.222 & 39.070 \\
            room0 & 37.812 & 38.037 & 38.059 & 38.057 & 38.063 & 38.032 & 38.104 & 38.095 & \textbf{38.160} & 38.027 & 38.030 & 37.891 & 37.948 \\
            room1 & 38.965 & 38.983 & 38.945 & 39.125 & 39.121 & 39.320 & 39.410 & 39.272 & 39.214 & 39.313 & \textbf{39.519} & 39.233 & 38.876 \\
            room2 & 39.642 & 39.846 & 39.981 & 39.964 & 40.009 & 39.918 & 39.928 & 39.972 & 40.040 & 39.991 & 40.053 & 40.143 & \textbf{40.364} \\
            \bottomrule
        \end{tabular}
    }
    \label{tab2}
\end{table*}

\begin{table*}[ht]
    \caption{Average Training Time and Model Size for Upsampling Methods on Mip-NeRF360 \cite{barron2022mipnerf360} and Replica \cite{replica19arxiv}.}
    \centering
    \scalebox{0.72}{
        \begin{tabular}{l|c|cccccc|cccccc|cccccc}
            \toprule 
            Metric & 3DGS & \multicolumn{6}{c|}{4x} & \multicolumn{6}{c|}{8x} & \multicolumn{6}{c}{16x}\\
            & & Lin. & Tri. & Spl. & MLS & Vor. & Dep. & Lin. & Tri. & Spl. & MLS & Vor. & Dep. & Lin. & Tri. & Spl. & MLS & Vor. & Dep. \\
            \midrule
            Mip-NeRF360 &  &  &  &  &  &  & &  &  &  &  &  &  &  &  &  &  &  &  \\
            \textbullet\ Time (s) & 1365 & 1365 & 1370 & 1409 & 1433 & 1519 & N/A & 1430 & 1438 & 1475 & 1503 & 1541 & N/A & 1575 & 1582 & 1620 & 1635 & 1657 & N/A \\
            \textbullet\ Size (MB) & 752 & 804 & 806 & 777 & 760 & 733 & N/A & 857 & 857 & 807 & 785 & 758 & N/A & 960 & 940 & 870 & 828 & 807 & N/A \\
            \midrule 
            Replica &  &  &  &  &  &  & &  &  &  &  &  &  &  &  &  &  &  &  \\
            \textbullet\ Time (s) & 487 & 502 & 506 & 523 & 540 & 569 & 714 & 549 & 536 & 559 & 568 & 605 & 830 & 607 & 606 & 627 & 634 & 683 & 930 \\
            \textbullet\ Size (MB) & 211 & 234 & 234 & 217 & 219 & 219 & 219 & 261 & 256 & 232 & 233 & 233 & 252 & 319 & 307 & 268 & 264 & 263 & 319 \\
            \bottomrule
        \end{tabular}
    }
    \label{tab3}
\end{table*}

\begin{figure}[t]
    \centering
    \begin{overpic}[width=0.47\textwidth]{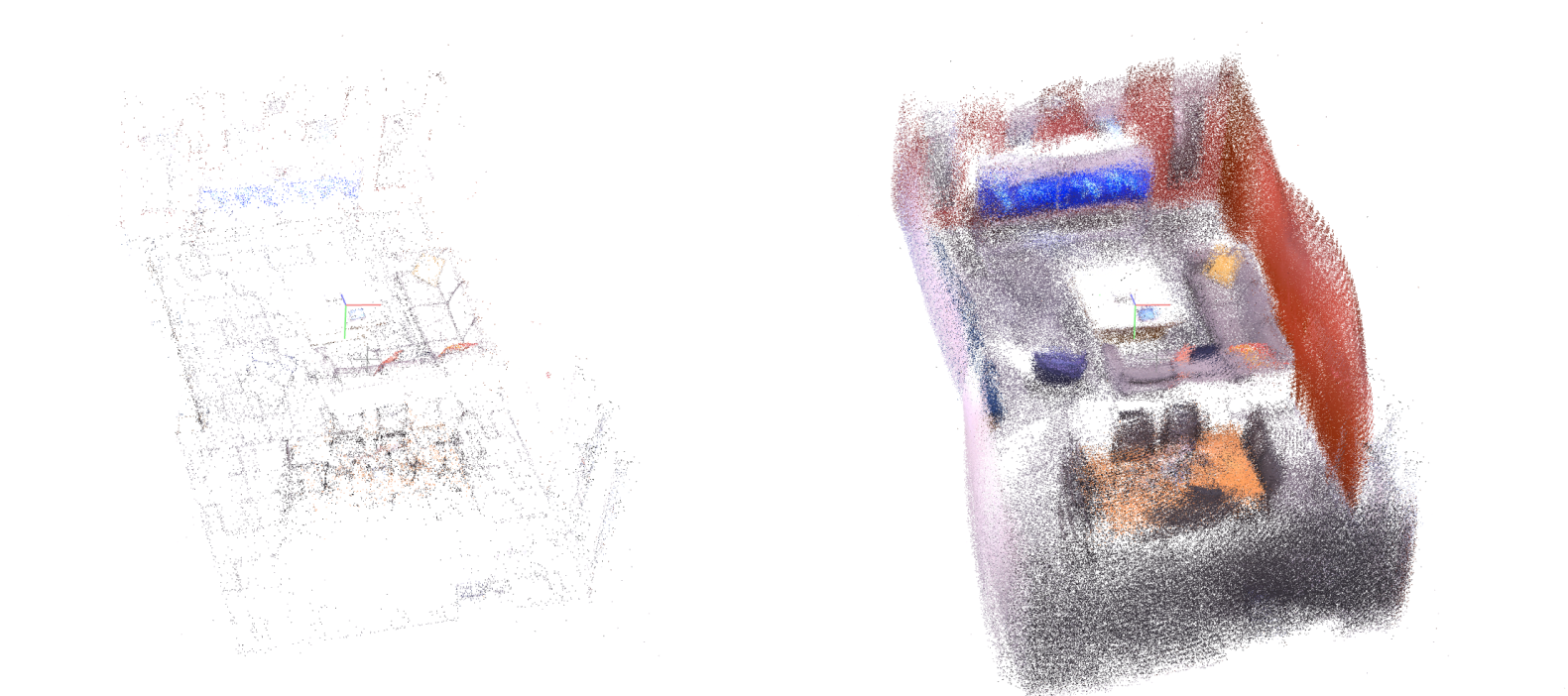}
        \put (20,43) {3DGS}
        \put (58,43) {Depth-guided}
    \end{overpic}
    \caption{The 32x upscaled point clouds of \textit{office3} from Replica \cite{replica19arxiv}.}
    \label{fig2}
\end{figure}

To further validate our methods and evaluate the depth-guided point lifting approach, we conduct additional experiments on Replica \cite{replica19arxiv}. A point cloud produced by the depth-guided approach is shown in Fig.~\ref{fig2}. Replica's \cite{replica19arxiv} scenes contains smaller initial points clouds (an average of 23000 points per scene compared to Mip-NeRF360's \cite{barron2022mipnerf360} 115000) so we investigate the effect of larger upsampling factors (16x and 32x). The results in Table~\ref{tab2} demonstrate that our methods generally improve the PSNR by an average or 0.646 db per scene.

Replica's \textit{office} and \textit{room} scenes contain similar objects and structure, resultantly we observe less variety in the methods with the highest PSNR. Most methods show improvement when scaling from 16x to 32x, suggesting more benefits from higher point densities. This contrasts with our findings on the Mip-NeRF360 \cite{barron2022mipnerf360} dataset, where benefits plateau earlier. The surface reconstruction methods demonstrate more stable performance across upsampling ratios, with consistent improvements at higher densities. This suggests their robustness in maintaining geometric fidelity even at higher upsampling rates.

\subsubsection{Guidelines for Method Selection}
Surface reconstruction methods (MLS and Voronoi) consistently outperform simple interpolation in all scenes. Again, as established with Mip-NeRF360 \cite{barron2022mipnerf360}, the Voronoi method improves initialization for scenes containing planar, low-texture surfaces. MLS maintains higher PSNR likely due to its ability to handle the mixture of planar surfaces (walls, desks) and complex objects. Also, since these scenes contain much less objects than scenes like \textit{counter}, so there are less opportunities for over-smoothing. The depth-guided point lifting approach shows continuous improvement at higher upsampling ratios and the most consistent performance across different scene types, suggesting that depth information provides valuable guidance for point placement. This method particularly excels in scenes with many low-texture objects at various depth levels, as evidenced by its performance in \textit{office1} and \textit{room2}. The performance of the depth-guided approach underscores the value of incorporating additional geometric information when available.

\subsection{Training Time and Model Size}
Analysis of computational results (Table~\ref{tab3}) reveals efficiency trade-offs across our methods. Training times scale sub-linearly with upsampling ratio, with the baseline 3DGS requiring 1365s on Mip-NeRF360 \cite{barron2022mipnerf360} and geometric methods showing increases of up to 20\% at 16x upsampling. The depth-guided approach demonstrates the highest computational cost on Replica \cite{replica19arxiv} (930s at 16x vs 487s baseline), reflecting the overhead of optimizing 3DGS with many promising points. Model size scales approximately linearly with upsampling ratio, though surface reconstruction methods show better efficiency than interpolation-based approaches. At 16x upsampling on Mip-NeRF360 \cite{barron2022mipnerf360}, while Linear interpolation required 960 MB (compared to 752 MB baseline), MLS and Voronoi maintain smaller footprints of 828 MB and 807 MB respectively. Notably, these trends are consistent across both datasets, with Replica showing proportionally similar scaling at lower absolute values due to reduced scene complexity. These findings suggest that while our methods introduce computational overhead, they maintain efficiency even at higher upsampling ratios, with surface reconstruction approaches offering the most favorable balance between quality improvement and resource utilization.

\subsection{Ablation Studies}
We conduct an ablation study comparing our methods against random point sampling within the scene bounding box. We generate points by uniformly sampling XYZ coordinates within the scene bounds and randomly assigning RGB values from the initial cloud. The results show significantly degraded performance compared to all proposed upsampling methods. In Mip-NeRF360 \cite{barron2022mipnerf360}, random sampling achieves PSNRs consistently 0.2 to 0.7 dB less than the original 3DGS initialization, especially at higher upsampling factors. This poor performance can be attributed to its failure to preserve local surface structure and color consistency, leading to noise in the initial Gaussian distribution that the optimization struggles to correct. This finding underscores that successful point cloud augmentation requires methods that respect the underlying scene geometry and maintain local color consistency, validating our approach of using geometry-aware upsampling strategies.

\section{Conclusion}
Our evaluations reveal that thoughtful augmentation of initial point clouds consistently improves reconstruction quality, with different methods showing distinct advantages based on scene characteristics. As a guideline, surface reconstruction methods excel in organic, detailed environments, while the Voronoi-based approach show strength in structured, planar scenes. Particularly, MLS demonstrates robust performance across varied environments, while simpler interpolation methods prove effective for scenes with many piecewise-smooth geometries. Our depth-guided approach shows particular promise to introduce scene wide details, particularly in low-texture regions that are missed by SfM. While computational overhead increases with upsampling ratio, the benefits to reconstruction quality often justify the additional resources. Future work might explore adaptive combinations of methods based on local scene geometry and traditional point cloud densification techniques could offer more performance gains. Learning-based approaches present an intriguing avenue, particularly given 3DGS's unique characteristic of concentrating Gaussians around high-frequency regions. 


\bibliographystyle{IEEEtran}
\bibliography{references.bib}

\end{document}